\documentclass{article}

\usepackage[authoryear,round,comma,sort&compress]{natbib}

\usepackage[preprint]{neurips_2025}

\usepackage[utf8]{inputenc} 
\usepackage[T1]{fontenc}    
\usepackage{hyperref}       
\usepackage{url}            
\usepackage{booktabs}       
\usepackage{amsfonts}       
\usepackage{amsmath}        
\usepackage{bm}             
\usepackage{microtype}      
\usepackage{xcolor}         
\usepackage{graphicx}       
\usepackage{algorithm}      
\usepackage{algorithmic}    

\hypersetup{
    colorlinks=true,
    linkcolor=black,
    citecolor=blue,
    urlcolor=blue,
    filecolor=blue,
    pdfborder={0 0 0}
}

\def\eqref#1{equation~\ref{#1}}

\def\1{\bm{1}}

\DeclareMathAlphabet{\mathsfit}{\encodingdefault}{\sfdefault}{m}{sl}
\SetMathAlphabet{\mathsfit}{bold}{\encodingdefault}{\sfdefault}{bx}{n}

\title{Building Interpretable Models for Moral Decision-Making}

\author{
Mayank Goel$^{1}$, Aritra Das$^{2}$, Paras Chopra$^{1}$ \\
$^{1}$Lossfunk \\
$^{2}$Ashoka University \\
\texttt{\{mayank.goel, paras\}@lossfunk.com, aritra.das@ashoka.edu}
}

\begin{document}

\thispagestyle{empty}
\begin{center}
\vspace*{-1cm}
\includegraphics[width=0.25\textwidth]{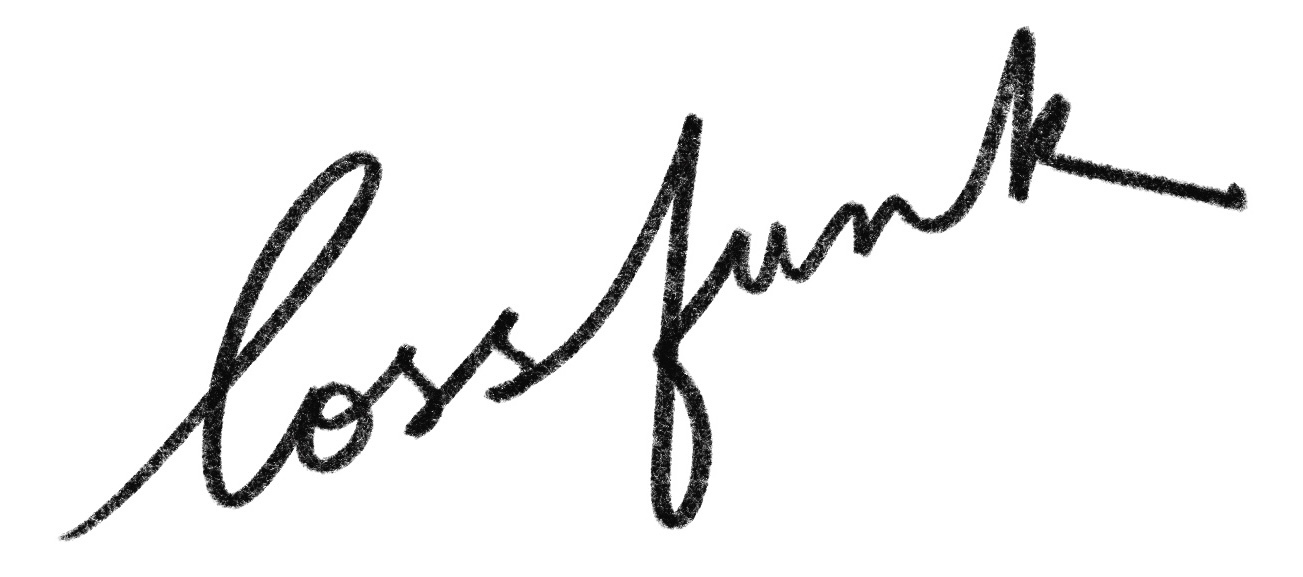}
\vspace{0.3cm}
\end{center}

\maketitle 

\begin{abstract}
We build a custom transformer model to study how neural networks make moral decisions on trolley-style dilemmas. The model processes structured scenarios using embeddings that encode who is affected, how many people, and which outcome they belong to. Our 2-layer architecture achieves 77\% accuracy on Moral Machine data while remaining small enough for detailed analysis. We use different interpretability techniques to uncover how moral reasoning distributes across the network, demonstrating that biases localize to distinct computational stages among other findings.  
\end{abstract}

\section{Introduction}

As AI systems increasingly confront moral dilemmas, from autonomous vehicles to content moderation, we face an urgent problem to understand not just what these systems decide but also how they reason and where moral biases originate within their architectures. We train a small model (104K Parameters) which can learn general moral principles and can be trained on ~3M scenarios from the Moral Machine Dataset \citep{Awad2018}. To address this challenge, we establish that moral principles must be encoded in the model architecture itself. Rather than probing opaque large language models post-hoc, we build a custom architecture from the ground up. Each token in our architecture represents three pieces of information: the character type, the number present in the scenario, and the team assignment. This representation encodes our hypothesis that moral computation reduces to identifying stakeholders, quantifying impact, and resolving competing interests. Further, we employ complementary interpretability methods to understand the model's moral reasoning process. We use causal intervention to quantify the influence of individual character types on decisions. Additionally, we apply layer-wise attribution to trace how different moral biases localize to specific computational stages across the network. Finally, we use circuit probing to identify sparse subnetworks that causally implement the final moral scores.
Overall, our work makes the following main contributions: 
\begin{itemize}
    \item Firstly, we design a 2-layer transformer specifically optimized for moral reasoning tasks, achieving 77\% accuracy on Moral Machine trolley problems. Our architecture demonstrates that transparency and predictive capability in moral choices are not mutually exclusive. 
    \item Secondly, we establish an exploratory interpretability analysis of moral decision-making combining causal intervention, layer-wise attribution, and circuit probing to map the internal mechanisms underlying moral judgment in transformers. 
\end{itemize}

The code is available at \href{https://github.com/Lossfunk/modeling-moral-machine}{GitHub}.

\begin{figure}[t]
\centering
\includegraphics[width=0.8\textwidth]{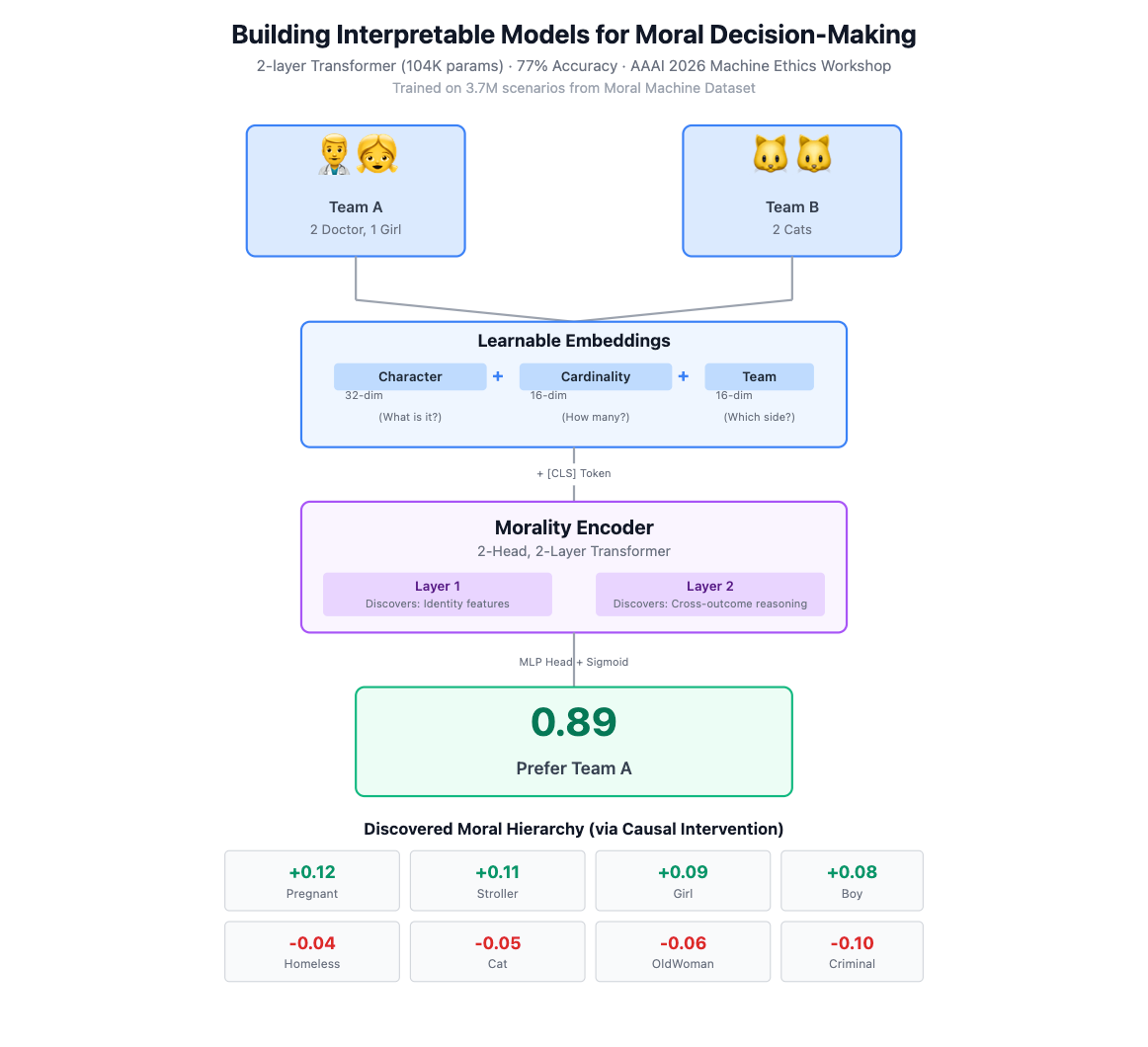}
\caption{Overview of the paper}
\label{fig:overview}
\end{figure}

\section{Related Work}
The Moral Machine dataset \citep{Awad2018} was originally a crowdsourced collection of ~40 million human decisions on AV dilemmas. Recently, it has been used to evaluate the ethical reasoning capabilities of large language models (LLMs). \citet{Jin2024} translated 1,000 trolley-problem vignettes into over 100 languages and compared LLM responses to cross-cultural human norms. \citet{Ahmad2025} evaluated 52 LLMs (GPT, Claude, Llama, etc.) on Moral Machine scenarios, finding larger models aligned more closely with human aggregate preferences. In parallel, several datasets have been developed to model moral and social norms. The ETHICS dataset \citep{Hendrycks2021} spans justice, virtue, duty, and commonsense, shows the partial success of Language Models at value-aligned judgment. Social Chemistry 101 \citep{Clark2020} provides 100K situations annotated with normative ``rules of thumb,'' and has been used to train models to generalize commonsense moral reasoning. Resources based on Moral Foundations Theory \citep{Graham2009} enable categorization of ethical text across key moral dimensions (e.g., care, fairness, loyalty). Mechanistic interpretability studies have begun to explore how transformer-based models encode moral reasoning, with work identifying distinct ``moral neurons'' activating on ethical stimuli \citep{Schacht2025} and applying PCA and probing techniques to isolate moral subspaces in representation space \citep{Schramowski2020}.

\section{Architecture}

We propose a transformer-based model designed specifically for moral reasoning over structured scenario representations. Unlike prior work that applies general-purpose LLMs to ethical dilemmas, our architecture exploits the compositional structure inherent in trolley-problem scenarios.

\subsection{Input Representation}

Each moral dilemma is represented as a pair of outcomes, where an outcome is a vector over 23 character types (e.g., Man, Woman, Criminal, Doctor). Given a scenario with outcomes $\mathcal{O}_0$ and $\mathcal{O}_1$, we encode each outcome as a sequence of tokens—one per character type. 

For every character $c$ with cardinality $n_c$ in outcome $\mathcal{O}_t$ (where $t \in \{0,1\}$ denotes the team, one or the other), we construct a compositional embedding:
\begin{equation}
\mathbf{e}_c^{(t)} = [\mathbf{E}_{\text{char}}(c) \,;\, \mathbf{E}_{\text{card}}(n_c) \,;\, \mathbf{E}_{\text{team}}(t)]
\end{equation}
where $[\cdot\,;\,\cdot]$ denotes concatenation, and the embedding dimensions are allocated as $d_{\text{char}} = d/2$, $d_{\text{card}} = d_{\text{team}} = d/4$, summing to total embedding dimension $d$.

This design choice is driven by our hypothesis that moral reasoning decomposes into: (1) \textit{who} is affected (character identity), (2) \textit{how many} are affected (cardinality), and (3) \textit{which side} they belong to (team membership).

\subsection{Transformer Processing}

We prepend a learnable \texttt{[CLS]} token to the sequence of 46 character tokens (23 per outcome) and process the full sequence with a standard transformer encoder. The architecture uses $L=2$ layers and $H=2$ attention heads with embedding dimension $d=64$. The relatively shallow architecture comes from the simplicity of our task itself: unlike natural language, our scenarios do not require deep hierarchical composition.

The transformer learns to compute cross-outcome comparisons via self-attention. Importantly, as we are not using position embeddings, the team embeddings allow the model to form separate representations of the two outcomes before computing their relative moral value through attention interactions.

\subsection{Classification Head}

The final \texttt{[CLS]} token representation is passed through a two-layer MLP with GELU activation to produce a scalar logit. This logit represents the model's preference for outcome 1 over outcome 0, which we convert to probabilities via sigmoid. The design mirrors recent findings in mechanistic interpretability \citep{Schacht2025} that moral reasoning in transformers localizes to specific aggregation points in the representation space. This probabilistic design naturally handles conflicts in the training set: contradictory preferences push the model toward intermediate probabilities, allowing it to learn which scenario features produce genuine moral uncertainty. This ensures that even with conflicting samples, our model ends up learning the uncertainty itself.

\subsection{Symmetry and Invariance}

A key property of moral judgments is \textit{side-invariance}: the choice between saving group A versus group B should be the complement of the choice between saving group B versus group A. To enforce this, we apply a symmetrization procedure at inference:
\begin{equation}
p(\mathcal{O}_1 \succ \mathcal{O}_0) = \frac{1}{2}\left[\sigma(f(\mathcal{O}_0, \mathcal{O}_1)) + \big(1 - \sigma(f(\mathcal{O}_1, \mathcal{O}_0))\big)\right]
\end{equation}
where $f$ is our model, $\sigma$ is the sigmoid function, and $\mathcal{O}_1 \succ \mathcal{O}_0$ denotes preferring outcome 1. This averages predictions from both orderings, guaranteeing consistent probability assignments regardless of input order. Another motivation for this is the lack of symmetry we see at inference time, motivating an invariant inference for better predictability.

\subsection{Model Configuration}

We explored several architecture configurations to balance representational capacity with model compactness. We use a subset of the Moral Machine data which had only people who filled out the survey form, which led to a dataset of 5.4M, of which 1.7M were unique scenarios and kept aside as the validation set, and the remaining were used as the training set. By using only unique situations for the validation, we avoid data leakage. Table~\ref{tab:config} reports validation accuracy on held-out Moral Machine scenarios across different choices of embedding dimension $d$, number of attention heads $H$, and transformer layers $L$.

\begin{table}[h]
\centering
\begin{tabular}{ccc|c}
\hline
\textbf{Embed Dim} & \textbf{Heads} & \textbf{Layers} & \textbf{Val Acc (\%)} \\
\hline
32 & 2 & 2 & 76.5 \\
64 & 2 & 2 & 77.1 \\
64 & 4 & 2 & 77.3 \\
64 & 4 & 3 & 77.5 \\
\hline
\end{tabular}
\caption{Validation accuracy across architecture configurations. We select $d=64$, $H=2$, $L=2$ as our final model.}
\label{tab:config}
\end{table}

While the configuration $d=64$, $H=4$, $L=3$ achieved the highest validation accuracy (77.5\%), we selected $d=64$, $H=2$, $L=2$ (77.1\%) for our final model to prioritize interpretability and computational efficiency. The marginal 0.4\% improvement did not justify the increased complexity for our mechanistic analysis.

The choice of $d=64$ over $d=32$ is motivated by our compositional embedding scheme. With $d=32$, character embeddings receive only $d_{\text{char}} = 16$ dimensions to encode 23 distinct character types with morally salient attributes (age, gender, profession, social status). This would have constrained the representational space available for capturing the moral distinctions central to our task. In contrast, $d=64$ allocates 32 dimensions to character identity, providing sufficient capacity while maintaining a compact model suitable for interpretability experiments.

\section{Interpretability Experiments}

We run a series of interpretability experiments specialized to take advantage of the transformers architecture and to help us understand morality. 

\subsection{Causal Intervention}

To measure which characters causally influence the model's decisions, we employ the DoWhy causal inference framework \citep{sharma2020dowhy}. We generate 20,000 synthetic moral scenarios and construct a causal model for each character $c$ with treatment $T_c \in \{0,1\}$ (whether $c$ appears in outcome 1), outcome variable $p(\mathcal{O}_1 \succ \mathcal{O}_0)$ (model's preference probability), and confounders $\text{total}_0, \text{total}_1$ (total individuals per outcome). Using backdoor adjustment with linear regression, we estimate the Average Treatment Effect:\\ $\text{ATE}_c = \mathbb{E}[\text{outcome} \mid T_c=1, \text{confounders}] - \mathbb{E}[\text{outcome} \mid T_c=0, \text{confounders}]$. This isolates the causal effect of $c$'s presence while controlling for group size confounding.

\begin{figure}[t]
\centering
\includegraphics[width=0.4\textwidth]{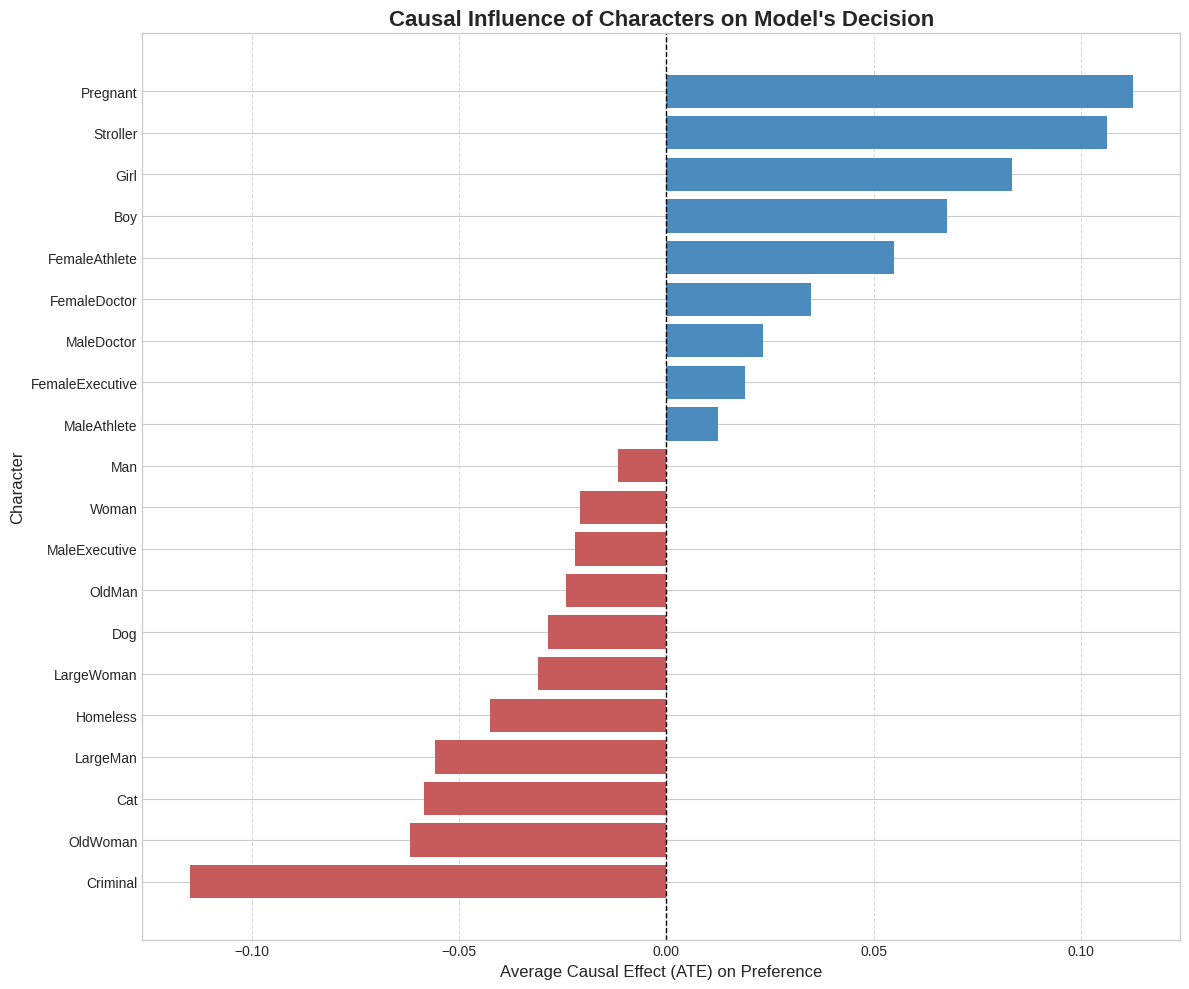}
\caption{Causal influence of character types on model decisions, estimated via DoWhy backdoor adjustment controlling for group sizes. Blue bars indicate positive causal effects (model favors outcomes containing these characters); red bars indicate negative effects.}
\label{fig:causal}
\end{figure}

Figure \ref{fig:causal} demonstrates a stark moral hierarchy. Pregnant (ATE = +0.12) and Stroller (ATE = +0.11) dominate positive influence, followed by Girl (+0.09) and Boy (+0.08). Conversely, Criminal shows the strongest negative effect (ATE = -0.10), with OldWoman (-0.06), Cat (-0.05), and Homeless (-0.04) also devalued. Generic categories (Man, Woman) cluster near zero (-0.01 to -0.02), suggesting they serve as moral baselines. The 22-percentage-point spread between Pregnant and Criminal indicates that character identity alone can shift preferences by over one-fifth of the probability space, demonstrating that the model has learned an implicit hierarchy of moral worth independent of utilitarian considerations.

\subsection{Layer-wise Bias Localization}

To identify where moral biases emerge in the network, we perform layer-wise attribution analysis by extracting attention weights from each transformer layer and correlating them with bias scores across five bias dimensions: legality (Criminal vs. law-abiding), gender (Man vs. Woman), social role (executives/doctors vs. homeless), age (children vs. elderly), and species (humans vs. animals). For each bias type, we generate contrastive scenarios and compute an importance score for each layer-head combination as $I_{\ell,h} = \text{Var}(\alpha_{\ell,h}) \cdot |\text{Corr}(\alpha_{\ell,h}, b)|$, where $\alpha_{\ell,h}$ are the attention weights from head $h$ in layer $\ell$, and $b$ is the bias score (preference for the privileged group). This metric captures both attentional selectivity (variance) and alignment with biased decisions (correlation).

\begin{figure}[t]
\centering
\includegraphics[width=0.5\textwidth]{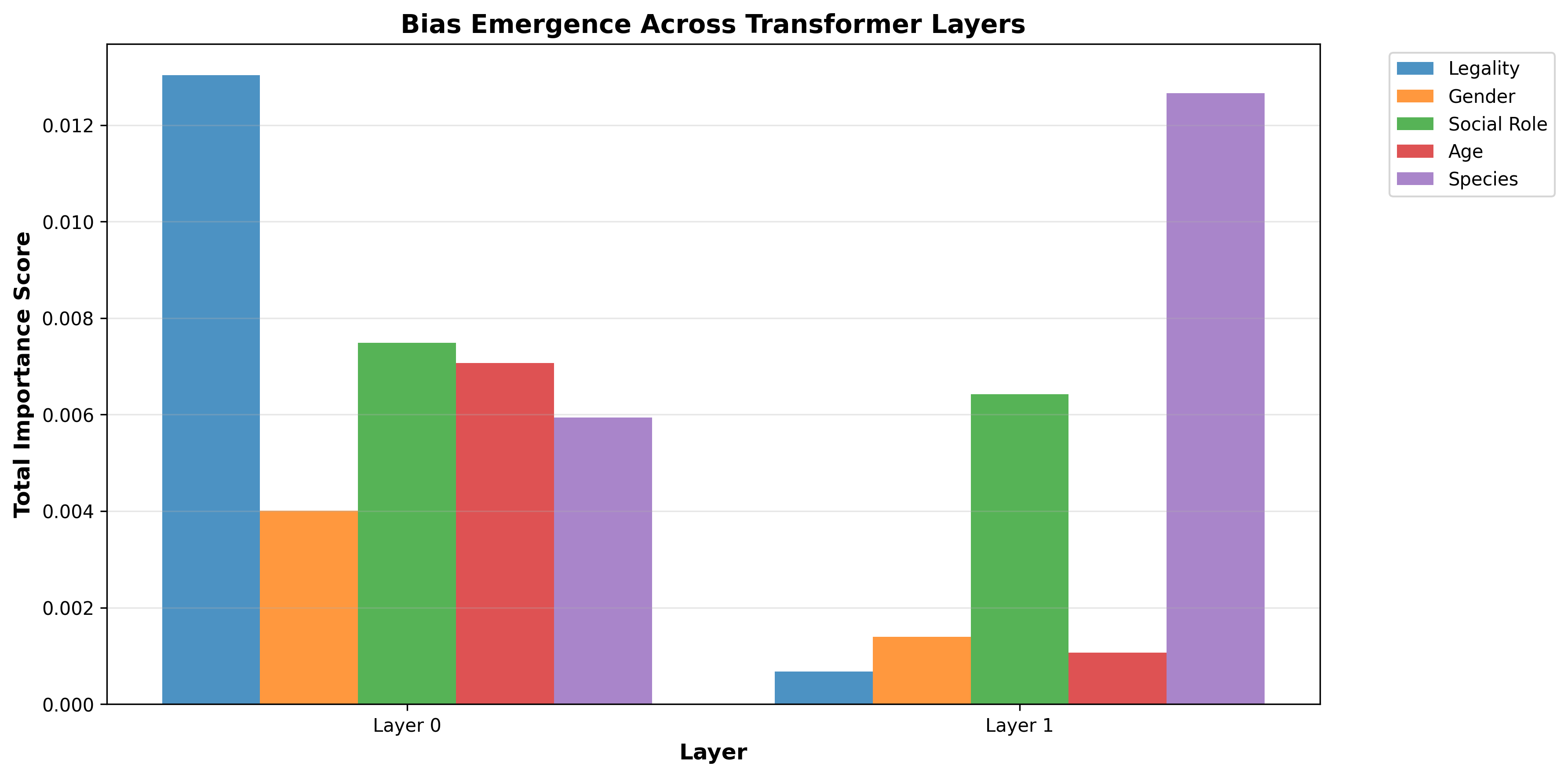}
\caption{Total importance scores for each bias type across transformer layers, computed by summing attention variance-correlation products across all heads. Layer 0 dominates legality bias while Layer 1 dominates species bias, indicating distinct computational stages.}
\label{fig:layer_emergence}
\end{figure}

\begin{figure}[t]
\centering
\includegraphics[width=0.5\textwidth]{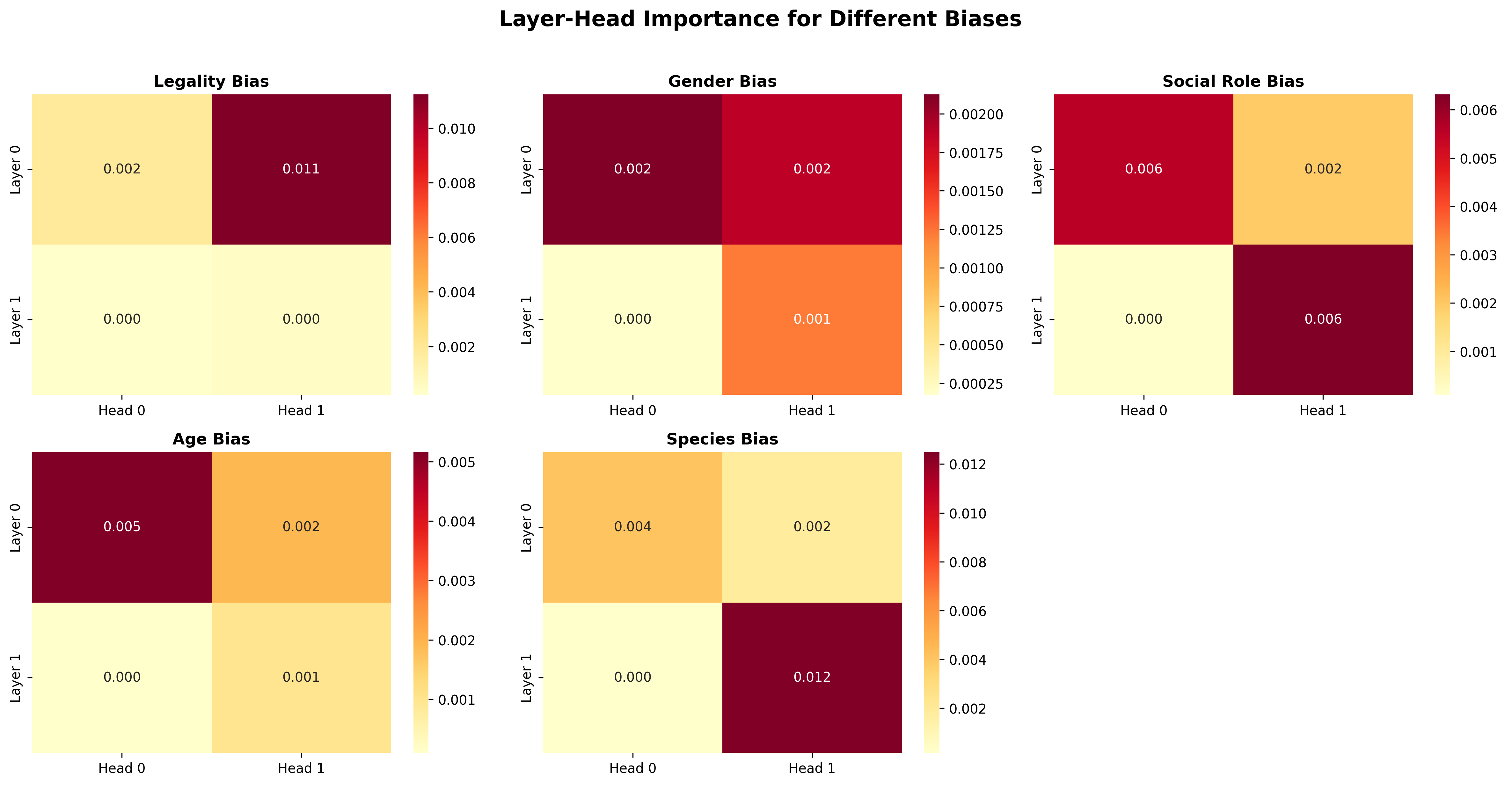}
\caption{Heatmaps showing importance scores for each layer-head combination across five bias dimensions. Dark red indicates high importance. Layer 0 Head 1 specializes in legality discrimination (0.011), while Layer 1 Head 1 specializes in species bias (0.012), revealing functional specialization despite the shallow architecture.}
\label{fig:layer_head}
\end{figure}

Figure \ref{fig:layer_emergence} shows that bias formation differs fundamentally across network depth. Legality bias localizes almost entirely to Layer 0 (importance = 0.013), with Layer 1 contributing negligibly. This suggests the model identifies criminal status early via character embeddings and carries this judgment forward. Conversely, species bias emerges predominantly in Layer 1 (importance = 0.013), indicating that human-animal distinctions require compositional reasoning over multiple characters. Age and social role biases distribute across both layers but concentrate in different heads (Figure \ref{fig:layer_head}), with Layer 0 Head 1 specializing in legality (0.011) while Layer 1 Head 1 specializes in species discrimination (0.012). This functional specialization suggests the shallow architecture nonetheless learns a division of labor, with early layers detecting salient individual attributes and later layers performing cross-outcome comparisons that privilege certain groups.

\subsection{Circuit Probing}
Beyond identifying where biases localize, we use circuit probing to discover which specific neurons causally implement the moral hierarchy we observed in causal intervention experiments. \citet{lepori2024uncovering} introduce circuit probing, which learns which neurons are responsible for computing specific intermediate variables by training sparse binary masks over a frozen model, then validates causality through targeted ablation while comparing against random subnetwork controls.

We adapt circuit probing to analyze our two-layer transformer encoder (embedding dimension 64, 2 attention heads per layer, MLP dimension 256) trained on the Moral Machine dataset. Our model uses compositional embeddings combining character identity, count, and team membership, processes sequences of the form $[\text{CLS} \mid \text{outcome}_0 \text{ tokens} \mid \text{outcome}_1 \text{ tokens}]$ through pre-norm transformer layers with \texttt{batch\_first=True}, and classifies via a CLS token through a two-layer MLP head ($64 \rightarrow 32 \rightarrow 1$ with GELU activation). Instead of probing for predefined linguistic variables, we extract per-character moral weights from the trained model using pairwise comparisons normalized to Man=1.0, then label each real scenario by the sign of the weighted score difference $\sum(\text{count}_{\text{left}} \times w) - \sum(\text{count}_{\text{right}} \times w)$ between outcomes. We train sparse neuron-level masks on both MLP blocks (gating the hidden vector after \texttt{linear1+activation} before \texttt{linear2}) and attention heads (gating concatenated head outputs before \texttt{out\_proj}) using the soft nearest-neighbors objective with $L_0$ regularization ($\lambda=1e\text{-}5$), continuous sparsification ($\beta$ annealed to 200), and class-balanced sampling to handle severe imbalance (22.3\% preferring right). We implement CLS-only gating to isolate interventions to the exact position used for classification, and evaluate via 1-nearest-neighbor accuracy on masked CLS updates plus behavioral accuracy drops on real test data after inverting the learned binary masks.

Probing layer 1's MLP block with 20,000 training examples achieved KNN accuracies of 0.956 (soft gate) and 0.951 (hard gate), indicating the CLS MLP update robustly encodes the model-derived scoring signal (Table \ref{tab:circuit_probing_results}). The discovered circuit was highly sparse, selecting only 45 of 256 neurons (17.6\%). Targeted ablation reduced agreement with model-score labels from 0.921 to 0.908 ($\Delta=0.012$) on 350,000 test examples. With baseline chance accuracy of 0.777 given class imbalance, the model's margin above chance was 0.144, making the circuit's causal contribution approximately 8.3\% of this margin. 

\begin{table}[htbp]
\centering
\caption{Circuit Probing Results for Layer 1 MLP Block}
\label{tab:circuit_probing_results}
\footnotesize
\begin{tabular}{lll}
\hline
\textbf{Metric} & \textbf{Value} & \textbf{Details} \\
\hline
\multicolumn{3}{c}{\textit{Probing Accuracy}} \\
KNN Accuracy (Soft) & 0.956 & 1-NN on CLS \\
KNN Accuracy (Hard) & 0.951 & Binary mask \\
\hline
\multicolumn{3}{c}{\textit{Circuit Properties}} \\
Selected Neurons & 45/256 & 17.6\% sparsity \\
Training Examples & 20,000 & 80\% split \\
Test Examples & 350,000 & 20\% split \\
\hline
\multicolumn{3}{c}{\textit{Causal Analysis}} \\
Full Model Acc. & 0.921 & Model-score \\
Ablated Acc. & 0.908 & Circuit removed \\
Ablation Drop & 0.012 & 1.2 pp \\
Baseline Chance & 0.777 & 22.3\% imbalance \\
Margin & 0.144 & 0.921 - 0.777 \\
Causal Share & 8.3\% & $\Delta$ / Margin \\
Random Control & $\approx$0 & Equal-sized \\
\hline
\end{tabular}
\end{table}

\subsection{Local Relevance}

To explain individual predictions at 
the token level, we compute gradient-weighted attention relevance for each character in a scenario. Following \citet{chefer2021transformerinterpretabilityattentionvisualization}, who propose a class-specific method for Transformers that computes gradient-weighted attention relevance per layer: $\bar{A}^{(b)} = I + E_h((\nabla A^{(b)} \odot R^{(n_b)})_+)$, then aggregates across blocks: $C = \bar{A}^{(1)} \cdot \bar{A}^{(2)} \cdot \ldots \cdot \bar{A}^{(B)}$, where the [CLS] row of $C$ provides token-level relevance scores.

We adapt this method to our moral reasoning model. Our implementation captures per-head attention weights and gradients through forward hooks on each encoder layer. For each layer, we compute the gradient of the output logit with respect to attention weights, derive $\nabla A$ via the chain rule through value projections, and construct $\bar{A}^{(b)} = I + E_h((\nabla A^{(b)} \odot A^{(b)})_+)$. Given our sequence structure [CLS] + team0(23) + team1(23), the [CLS] row of $C$ yields 46 relevance scores. We normalize these to sum to 1.0 and map token indices to character names. To account for our model's symmetry property, we compute explanations for both original and team-swapped scenarios, then average the remapped scores. The output provides character-level relevance scores indicating each character's contribution to the decision. For the scenario \texttt{\{'Man': 3\}} vs \texttt{\{'Criminal': 3\}}, we notice that the model's decision is primarily driven by the ``Criminal'' token (relevance score: 0.27), which accounts for approximately 27\% of the total positive evidence favoring Team 1. Context tokens such as CrossingSignal (0.04) and Intervention (0.04), along with medical professions (FemaleDoctor: 0.03, MaleDoctor: 0.03), provide secondary support. Notably, demographic tokens like ``Man'' contribute relatively weak evidence (0.01 on Team 0, 0.01 on Team 1) compared to role-based and contextual features. 

\section{Discussion}

Our 2-layer transformer achieves 77\% accuracy on Moral Machine scenarios while remaining tractable for mechanistic analysis demonstrating that moral competence does not require large pretrained models. A simple model trained on the intuitive notion of what a trolley problem can be structured as is also a valid way of exploring morality.

The interpretability experiments show multiple interesting things about morality as learnt through the dataset - pointing out that the human notions of morality themselves can be learnt through training models on the data.  
The approach has clear limitations: training on aggregate human preferences inherits cultural biases. However, transparency enables new intervention strategies. Knowing criminal bias localizes to Layer 0 Head 1 allows targeted debiasing orthogonalizing representations or clamping attention weights—rather than coarse dataset rebalancing. We hope to extend this this line of work to traditional LLMs on moral questions. Future work along this direction will attempt to use this work as a base to explore larger LLMs on moral questions. 

\section{Acknowledgment}

We acknowledge Lossfunk for hosting this research and providing compute support.

\label{sec:references}

\bibliographystyle{apalike}
\bibliography{references}


\end{document}